%% file: main.tex
\documentclass[sigconf]{acmart}
\AtBeginDocument{%
  }

\copyrightyear{2023}
\acmYear{2023}
\setcopyright{rightsretained}
\acmConference[RecSys '23]{Seventeenth ACM Conference on Recommender
Systems}{September 18--22, 2023}{Singapore, Singapore}
\acmBooktitle{Seventeenth ACM Conference on Recommender Systems (RecSys
'23), September 18--22, 2023, Singapore,
Singapore}\acmDOI{10.1145/3604915.3608792}
\acmISBN{979-8-4007-0241-9/23/09}

\input{math-and-symbols.tex}

\usepackage[utf8]{inputenc} % allow utf-8 input
\usepackage[T1]{fontenc}    % use 8-bit T1 fonts
\usepackage{amsfonts}       % blackboard math symbols

\usepackage{multirow}
\usepackage{bm}
\usepackage{algorithm,algorithmic}
\usepackage{subcaption}
\usepackage{wrapfig}

\begin{document}

\title{Online Matching: A Real-time Bandit System for Large-scale Recommendations}

\author{Xinyang Yi}
\email{xinyang@google.com}
\affiliation{
  \institution{Google Deepmind}
  \city{Mountain View}
  \state{California}
  \country{USA}
}

\author{Shao-Chuan Wang}
\email{scwang@google.com}
\affiliation{
  \institution{Google Inc}
  \city{Mountain View}
  \state{California}
  \country{USA}
}

\author{Ruining He}
\email{ruininghe@google.com}
\affiliation{
  \institution{Google Deepmind}
  \city{Mountain View}
  \state{California}
  \country{USA}
}

\author{Hariharan Chandrasekaran}
\email{hariharan@google.com}
\affiliation{
  \institution{Google Inc}
  \city{Mountain View}
  \state{California}
  \country{USA}
}

\author{Charles Wu}
\email{charleswu@google.com}
\affiliation{
  \institution{Google Inc}
  \city{Mountain View}
  \state{California}
  \country{USA}
}

\author{Lukasz Heldt}
\email{heldt@google.com}
\affiliation{
  \institution{Google Inc}
  \city{Mountain View}
  \state{California}
  \country{USA}
}

\author{Lichan Hong}
\email{lichan@google.com}
\affiliation{
  \institution{Google Deepmind}
  \city{Mountain View}
  \state{California}
  \country{USA}
}

\author{Minmin Chen}
\email{minminc@google.com}
\affiliation{
  \institution{Google Deepmind}
  \city{Mountain View}
  \state{California}
  \country{USA}
}

\author{Ed H. Chi}
\email{edchi@google.com}
\affiliation{
  \institution{Google Deepmind}
  \city{Mountain View}
  \state{California}
  \country{USA}
}

\renewcommand{\shortauthors}{Yi et al.}
%%
%% Article type: Research, Review, Discussion, Invited or position
% \acmArticleType{Review}
%%
%% Links to code and data
% \acmCodeLink{https://github.com/borisveytsman/acmart}
% \acmDataLink{htps://zenodo.org/link}
%%
%%
%% Sometimes the addresses are too long to fit on the page.  In this
%% case uncomment the lines below and fill them accodingly.
%%
%% \authorsaddresses{Corresponding author: Ben Trovato,
%% \href{mailto:trovato@corporation.com}{trovato@corporation.com};
%% Institute for Clarity in Documentation, P.O. Box 1212, Dublin,
%% Ohio, USA, 43017-6221}

\begin{abstract}
The last decade has witnessed many successes of deep learning-based models for industry-scale recommender systems. These models are typically trained offline in a batch manner. While being effective in capturing users' past interactions with recommendation platforms, batch learning suffers from long model-update latency and is vulnerable to system biases, making it hard to adapt to distribution shift and explore new items or user interests. Although online learning-based approaches (e.g., multi-armed bandits) have demonstrated promising theoretical results in tackling these challenges, their practical real-time implementation in large-scale recommender systems remains limited. First, the scalability of online approaches in servicing a massive online traffic while ensuring timely updates of bandit parameters poses a significant challenge. Additionally, exploring uncertainty in recommender systems can easily result in unfavorable user experience, highlighting the need for devising intricate strategies that effectively balance the trade-off between exploitation and exploration. In this paper, we introduce \textsl{Online Matching}: a scalable closed-loop bandit system learning from users' direct feedback on items in real time. We present a hybrid \textsl{offline + online} approach for constructing this system, accompanied by a comprehensive exposition of the end-to-end system architecture. We propose Diag-LinUCB -- a novel extension of the LinUCB algorithm -- to enable distributed updates of bandits parameter in a scalable and timely manner. We conduct live experiments in YouTube and show that Online Matching is able to enhance the capabilities of fresh content discovery and item exploration in the present platform.
\end{abstract}

\maketitle

\input{introduction.tex}
\input{related_work.tex}
\input{methods.tex}
\input{system_overview.tex}

\input{results.tex}
\input{conclusion.tex}

\bibliographystyle{ACM-Reference-Format}
\bibliography{main.bib}

\end{document}

%% file: math-and-symbols.tex
\usepackage{amsmath,amsfonts,bm}

\def\eqref#1{(\ref{#1})}
\def\1{\bm{1}}

\def\rvb{{\mathbf{b}}}
\def\rvc{{\mathbf{c}}}
\def\rvd{{\mathbf{d}}}

\def\rvu{{\mathbf{i}}}

\def\rvu{{\mathbf{u}}}
\def\rvv{{\mathbf{v}}}
\def\rvw{{\mathbf{w}}}
\def\rvx{{\mathbf{x}}}
\def\rvy{{\mathbf{y}}}

\def\rmA{{\mathbf{A}}}

\def\rmI{{\mathbf{I}}}

\def\vtheta{{\bm{\theta}}}

\DeclareMathAlphabet{\mathsfit}{\encodingdefault}{\sfdefault}{m}{sl}
\SetMathAlphabet{\mathsfit}{bold}{\encodingdefault}{\sfdefault}{bx}{n}

%% file: introduction.tex
\section{Introduction}

Recommender systems have become indispensable for users to explore and access content in the era of information overload. Matching users with items to fulfill their information need by leveraging user and item data is a fundamental task in recommender systems.
Compared to other machine learning domains such as language and vision where data is static,  recommender systems are distinguished by dynamic user-system interactions that can give rise to a feedback loop, because new system policies are trained primarily based on data generated from users' past interactions. This feedback loop can further lead to the \textsl{rich gets richer} problem, i.e., future recommendations may excessively prioritize items with high engagement in the past, thus emphasizing the need for exploration. Furthermore, real-world applications must be capable of processing substantial amounts of fresh content. For instance, millions of new videos are uploaded to YouTube on a daily basis. As users demand up-to-date information, it is important for recommender systems to explore their preferences on new items and promptly incorporate their feedback into the decision-making processes. 

In recent years, deep learning-based models have emerged as a dominant class of approaches for building recommender systems. Notably, the majority of such models still rely on supervised learning from users' past interactions in \textsl{batch} mode, and their ability to explore and adapt in real-time is limited, if at all. In recent years, a line of work (e.g.~\cite{Gilotte2018, Adith2015, Philip2016, Chen2022}) applied offline reinforcement learning (RL) for deep learning-based models to address system bias in the training data. Nonetheless, such offline RL approaches suffer from large latency in model updates, since they still rely on learning from logged user feedback in the \textsl{batch} training framework, rather than through real-time interactions with the environment. As a result, offline RL is typically inadequate to explore new items in a timely and efficient manner, or to respond quickly to system changes such as shifts in user preferences.

In this paper, we focus on building a system with the capabilities of real-time learning and exploration. While online learning (e.g., multi-armed bandits) is well suited for the aforementioned considerations, the applications of it to large-scale real-world systems are much less explored compared to batch modeling. Practically, the challenges are mainly three-folds:

\begin{itemize}
\item Large exploration space. The exploration space typically grows linearly with the number of candidate items in the system, which poses a significant challenge for most real-world systems that need to retrieve from millions of items. In the absence of an appropriately designed space pruning technique, exploration can come at a high cost to the user experience.

\item Limited exploration traffic. Compared to traditional recommender systems that focus on \textsl{exploiting} known learnings, exploration by its nature can lead to a short-term user experience degradation/regret. Therefore it is crucial to perform online exploration in a \textsl{cost-efficient} manner, typically within a limited budget such as 1\% of randomly shuffled daily user traffic.

\item Lack of real-time learning systems. Existing systems are mainly designed for training models in batch mode and serve them online with fixed parameters. It is certainly non-trivial to build a custom highly scalable system for real-time learning with large data throughput.
\end{itemize}

This paper aims to tackle these challenges by introducing a real-time bandit system for item exploration by learning users' direct feedback on items in a closed loop. The proposed system, called Online Matching, adopts a hybrid \textsl{offline and online} learning approach in order to make the exploration system efficient, scalable, and responsive.  The offline learning component of Online Matching is designed for pruning the large exploration space and enabling cost-efficient exploration. 
In particular, we train a two-tower neural network model to co-embed users and items into the same space, similar to some existing neural retrieval systems such as \cite{yi2019samplingbias}. Then we discretize the embedding space into a specified number of ``user clusters'' using off-the-shelf embedding clustering approaches. Finally, we build a sparse bipartite graph between user clusters and the items to be explored based on their embedding similarities.
The key insight is that for a particular user cluster, items nearby in the embedding space can be good candidates to explore.
Note that all the aforementioned steps are performed offline. User cluster-item edges in the sparse bipartite graph essentially encode most promising user cluster-item pairs to explore. The online learning component of Online Matching is based on our novel extension of the classic LinUCB algorithm \cite{pmlr-v15-chu11a}, called Diag-LinUCB, with a user's real-time distribution over all the user clusters as the \textsl{context}.
With this novel algorithmic design, our bandit parameters simply correspond to user cluster-item edges in the sparse bipartite graph and can be updated in real-time in a fully distributed manner. The resulting online system boasts a very low policy update latency, which refers to the amount of time it takes from user interaction to incorporating the feedback in the decision-making process.

Online Matching has been successfully deployed to YouTube for the use cases of fresh content discovery and item exploration. In the first case, our goal is to quickly identify high-quality fresh items through exploration with small traffic, and amplify their impact through exploitation with major traffic. For item exploration, we aim to substantially increase discoverable corpus by leveraging large exploration traffic while having minimum regret on user experience. We provide the experimental frameworks for running these two experiments, and demonstrate the effectiveness of Online Matching through improved topline metrics in these two cases.

In a nutshell, our contributions are:
\begin{itemize}
\item \textbf{Algorithmic methods.} We provide a novel algorithmic framework that combines offline batch learning and online learning for building large-scale real-time bandit systems. Particularly, we propose Diag-LinUCB, a novel bandit algorithm that enables distributed bandit updates, allowing the system to scale and serve billions of users and millions of items.
\item \textbf{System design.} We present a complete system design, starting from the offline learning pipeline for creating a sparse bipartite graph, to the real-time bandit parameters aggregation performed by an online agent.
\item \textbf{Experiment design.} We study two use cases of Online Matching -- \textsl{Fresh Content Discovery} (Type-I) and \textsl{Corpus Exploration} (Type-II), and present the experiment frameworks for impact measurements in both cases. This includes the application of a user-corpus partition framework to measure the growth of the discoverable item corpus.
\item \textbf{Live experiments.} We conduct live experiments for both Type-I and Type-II use cases in YouTube with significant top-line metric gains. We demonstrate the value of Online Matching in improving user experiences through discovering high-quality fresh videos and growing discoverable item corpus.
\end{itemize}

%% file: related_work.tex
\section{Related Work}

\subsection{Neural Recommenders and Off-policy Learning}
Neural modeling has emerged as the dominant approach for constructing large-scale recommender systems in a range of industry applications, including video discovery \cite{paul2016}, news recommendations \cite{Okura2017EmbeddingbasedNR}, and social networks \cite{Liu2017, Zhai2017}. These models cover a broad range of tasks, from retrieval \cite{yi2019samplingbias} to ranking \cite{heng16, zhe19watchnext}.
The main focus of this paper is on the retrieval problem, where the goal is to find a few related items from a large item corpus. This problem is also known as \textsl{deep retrieval} and has been extensively studied in the past. For example, a line of work \cite{yi2019samplingbias, yang2020, pre-training, Yao2021} studied two-tower models for learning user and item representations in the same embedding space, allowing converting the problem of retrieval to maximum-inner-product-search (MIPS) \cite{scam_paper} with sub-linear complexity. Compared to matrix factorization and extreme classification models \cite{paul2016}, two-tower architecture can leverage content features of items, making it more suitable for generating reasonable representations of fresh items with little user engagement. Although a large amount of training data is often available in real-world applications, neural models (including aforementioned two-tower models) are typically trained offline on logged user feedback in a supervised fashion, making them biased toward existing recommendation policies and hard to promptly adapt to system changes.
In this paper, we propose an efficient exploration space pruning strategy, using two-tower models as a key part of our offline learning framework. Benefiting from the generalization capabilities of two-tower models, our strategy makes online learning feasible under strict latency and traffic constraint. 

To alleviate the aforementioned bias problem, there has been a body of work bringing offline reinforcement learning (RL) techniques to neural recommender systems. The idea of off-policy learning is to estimate the value of a target policy or to train it using data collected by a different behavior policy. A series of work (see e.g. \cite{Gilotte2018, Adith2015, Philip2016}) focused on developing off-policy estimators through inverse-propensity weighting. For target policy learning, Chen et al. \cite{Chen2022} applied off-policy correction to training a neural sequence model for video retrieval. There are also various papers (see e.g. \cite{Thorsten2017, zhe19watchnext}) studying learning unbiased ranking models from biased logged feedback. More recently, Ma et al. \cite{Ma2020} proposed using off-policy learning for two-stage recommender systems. One challenge in off-policy learning is that data collected from the existing behavior policy might not accurately represent the target policy. This is especially true for fresh items with minimal or no engagement data, where the inverse propensity weighting can have a high variance.
Offline RL, like traditional batch trained neural recommenders, is also subject to significant delays in data logging, model training, and deployment, as it is still based on the batch training paradigm.

\subsection{Online Learning and Exploration}
Online learning provides a useful framework for building recommenders that are adaptable to user feedback. One commonly used type of online learning algorithm is bandit algorithms, such as UCB \cite{auer2002}, Thompson Sampling \cite{Chapelle2011}, and LinUCB \cite{pmlr-v15-chu11a}. These algorithms are designed to balance the exploration of new options with the exploitation of known good options, allowing the system to learn and adapt to user behavior over time.
The application of bandits to recommenders dates back to the seminal work by Li et al. \cite{Li_2010} on using contextual bandits for news recommendation. The work in \cite{Chapelle2011} provided an evaluation of Thompson Sampling on real-world datasets.  Jeremie et al. \cite{Jeremie2015} studied bridging matrix factorization and bandits to solve the cold-start problem in recommenders. Besides the use of bandits for exploring new items, more recently, Song et al. \cite{Song2022} proposed creating a hierarchical representation of item space to help explore new user interests. To the best of our knowledge, most of the works on bandits for recommendations conduct experiments on offline synthetic or real-world datasets with simulated environment, and how to scale online learning system to billions of users and millions of items with real-time updates is not well studied. In contrast, a key contribution of this paper is on scaling in real-world environment, and we provide the engineering details on how we build the Online Matching system. 

\begin{figure}
  \centering
  \includegraphics[width=0.45\textwidth]{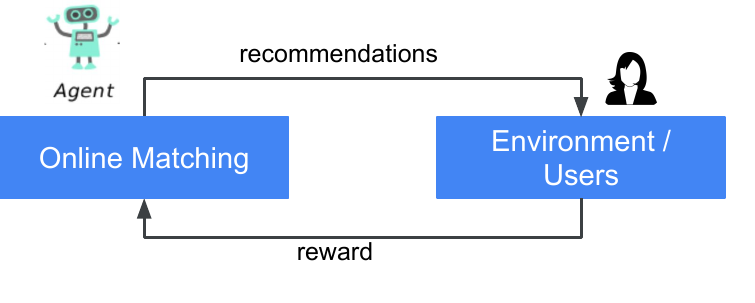}
  \caption{Illustration of how Online Matching interacts with users and environment.}
  \label{fig:illustration-om}
  \Description[A diagram of Online Matching interacting with users and environment.]{The Online Matching agent makes recommendations to the users and environment and observes rewards.}
\end{figure}

\subsection{Real-time Recommenders}
Building real-time systems that can quickly process user feedback is critical for recommenders, especially in applications where new items are quickly added. StreamRec \cite{chandramouli2011streamrec} is a demo for event-driven fast data processing for training recommender models. Recently, Monolith was introduced in \cite{liu2022monolith} as a system for fast model training with collisionless embedding table. In addition to the real-time update capability, our Online Matching system is further built with explicit exploration strategies to address the feedback loop problem \cite{Chaney2018} that is not addressed in this stream of work.

%% file: methods.tex
\section{Methods}

\subsection{Background}

We aim to build Online Matching by learning users' feedback to items in a closed loop, see Fig. \ref{fig:illustration-om} for an illustration. One naive approach is to apply multi-armed bandits for each user. That is for each user, we maintain a set of bandit parameters for all the explored items.
This approach is not scalable because when there is a large number of explored items, many trials are needed for each user, making it hard for bandit parameters to converge, especially in the case new items are continuously added. Therefore, we need to consider contextual bandits where user context is modeled to allow using context features and across-user learning. Formally, let $t$ denote the current time step, $a_t \in \mathcal{A}$ denote the action selected at time $t$ from the entire action space $\mathcal{A}$, and $r_t(a_t)$ denote the reward obtained by selecting action $a_t$ at time $t$. The goal is to learn a policy $\pi_t(\rvx_t)$ that maps the observed user feature $\rvx_t \in \mathbb{R}^d$ to an action $a_t$, such that the expected cumulative reward is maximized over a sequence of $T$ time steps:

\begin{equation}
\max_{\pi_t} \mathbb{E}\left[\sum_{t=1}^T r_t(a_t)\right]
\end{equation}

In contextual linear bandits, the reward function is assumed to be linear in the feature vector $\rvx_t$, i.e.,
\begin{equation} \label{bandit_reward}
\mathbb{E}\left[r_t(a)\right] = \rvx_t^\top \vtheta_a^*,
\end{equation}
for all $t$, where $\vtheta_a^*$ is an \textsl{unknown} weight vector associated with action $a$. The goal is to estimate the weight vectors $\vtheta_a^*$ ($a \in \mathcal{A}$) and use them to select actions that maximize the expected reward. A more generic formulation \cite{pmlr-v15-chu11a} is assuming there is a single unknown weight vector $\vtheta^*$, and the expectation reward is $\rvx_{t,a}^\top \vtheta^*$, where $\rvx_{t,a}$ represents both user and item features. We choose not to model item features in this work because we find item id is a very important feature to reflect the finest difference among various items, and adding this feature to $\rvx_{t,a}$ can make this vector to be very high-dimensional. The setup in \eqref{bandit_reward} is also called disjoint linear models as studied in a few papers \cite{Li_2010, deshpande2012, wang2017}.

The LinUCB algorithm \cite{Li_2010} maintains an estimate of the unknown parameter vector $\vtheta_{a, t}$ for each action $a$ at each time step $t$. Let $\rmA_{a,t}$ and $\rvb_{a,t}$ be the $d$-by-$d$ positive definite matrix and $d$-dimensional vector that represent the covariance matrix and mean vector of the context features up to time $t$, respectively. Then the estimate of $\vtheta_{a, t}$ is given by:
\begin{equation}\label{eq:linucb_update_theta}
\vtheta_{a,t} = \rmA_{a,t}^{-1} \rvb_{a,t}.
\end{equation}
LinUCB selects the arm with the highest upper confidence bound, namely:
\begin{equation}\label{eq:linucb_ucb}
UCB_a(t) = \rvx_t^T \vtheta_{a, t} + \alpha \sqrt{\rvx_{t}^T \rmA_{a,t}^{-1} \rvx_t},
\end{equation}
where $\alpha$ is a hyperparameter that controls the exploration-exploitation tradeoff. If action $a$ is chosen, $\rmA_{a,t}$ and $\rvb_{a,t}$ are then updated using the observed reward $r_{a,t}$ and context vector $\rvx_t$ through:
\begin{equation} \label{eq:linucb_update}
\rmA_{a,t} \leftarrow \rmA_{a, t-1} + \rvx_t \rvx_t^T,~ \rvb_{a, t} \leftarrow \rvb_{a, t-1} + \rvx_t r_{a,t},
\end{equation}
For any action $a$ not chosen, we have $\rmA_{a,t} \leftarrow \rmA_{a, t-1}, \rvb_{a, t} \leftarrow \rvb_{a, t-1}$. See Algorithm \ref{alg:linucb} for a detailed description.

\begin{algorithm}[H]
\caption{LinUCB Algorithm \cite{Li_2010}}
\label{alg:linucb}
\begin{algorithmic}[1]
\STATE \textbf{Input}: context vector $\rvx_t$ for each time step $t$, number of arms $N$, hyperparameter $\alpha$,
\STATE Initialize $\rmA_{a,0} = \rmI_d, \rvb_{a,0} = \mathbf{0}_d$ for all arms $a \in [N]$.
\FOR{t = 1 to T}
\STATE Observe context vector $\rvx_t$.
\FOR{j = 1 to N}
\STATE Compute $UCB_j(t)$ using current estimate of $\vtheta_j$ (Eq. \eqref{eq:linucb_ucb} ).
\ENDFOR
\STATE Select arm $a_t$ with the highest $UCB_a(t)$.
\STATE Observe reward $r_{a_t,t}$.
\STATE Update $\rmA_{a_t,t}$ and $\rvb_{a_t,t}$ using Eq. \eqref{eq:linucb_update}.
\STATE Update $\vtheta_{a_t,t}$ using Eq. \eqref{eq:linucb_update_theta}.
\STATE For $a \neq a_t$, $\rmA_{a,t} \leftarrow \rmA_{a, t-1}, \rvb_{a, t} \leftarrow \rvb_{a, t-1}$.
\ENDFOR
\end{algorithmic}
\end{algorithm}

\textbf{Scaling problems of LinUCB}. There are several practical issues that prevent us from directly using LinUCB for serving online traffic and getting real-time updates:
\begin{itemize}
\item \textbf{Large Action Space}. The computation of $UCB_j(t)$ is over all actions or items. In our application, even after narrowing the corpus down to fresh items, there are still millions of items worth exploring.
\item \textbf{Covariance Inversion}. Computing $UCB_j(t)$ and updating $\theta_{a,t}$ depend on calculating the inverse of the covariance matrix $\rmA_{a,t}$ that has a size of $d$-by-$d$. Note that the covariance matrix cannot be precomputed or cached due to the streaming updates. The online computational cost could be prohibitively high when facing a large user traffic.
\item \textbf{Synchronous Updates.} When one item is exposed to users, there could be many feedback received from various users. The streaming updates of $\rmA_{a,t}$ and $\rvb_{a,t}$ require an item-level synchronization, which poses a challenge on maintaining a high throughput to handle large traffic. As LinUCB converges to exploiting fewer top items, synchronization cost could be a real bottleneck.
\end{itemize}
\subsection{Sparse bipartite graph} \label{sec:sparse_bipartite_graph}
\begin{figure}[ht]
  \centering
  \begin{subfigure}[b]{0.36\textwidth}
    \centering
    \includegraphics[width=0.6\textwidth]{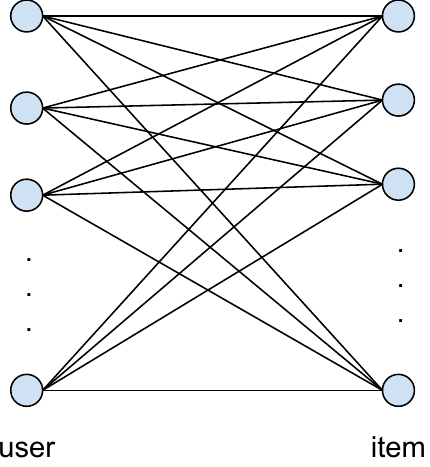}
    \caption{}
    \label{fig:ui-dense}
  \end{subfigure}
  \hfill
  \begin{subfigure}[b]{0.52\textwidth}
    \centering
    \includegraphics[width=0.6\textwidth]{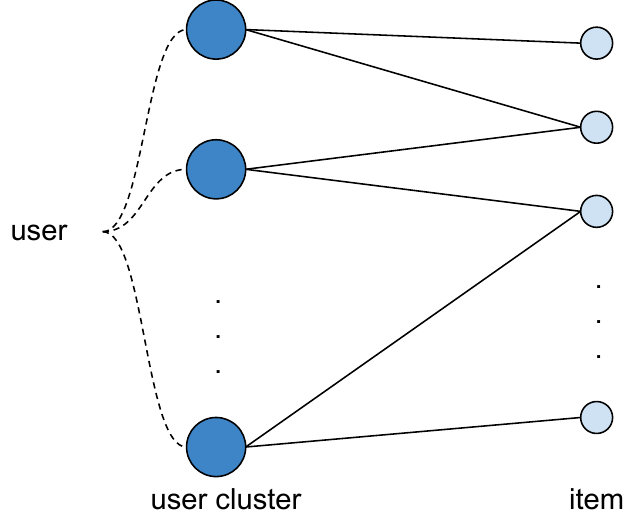}
    \caption{}
    \label{fig:ui-sparse}
  \end{subfigure}
  \caption{Illustration of the user-item matching problem: (a) Dense bipartite graph between users and items; (b) Sparse bipartite graph between user clusters and items.}
  \label{fig:ui}
  \Description[Comparison of dense and sparse bipartite graph.]{Our bipartite graph is based on user clusters and the links from user clusters to items are much sparser.}
\end{figure}

To overcome the aforementioned challenges, we propose a novel variant of LinUCB. Before diving into that, we take a step back to build some intuitions for our solution. Let's look at the problem of user-item matching from a graph perspective. As shown in Fig. \eqref{fig:ui-dense}, suppose there is a dense bipartite graph between users and items, the main goal of recommendation is to find good edges in the graph to make the connections between users and items. Naively, we could explore all items for each user disjointly, but apparently this would not be efficient. Our idea is to group users by clusters, and then reduce the exploration space in each cluster. As illustrated in Fig. \eqref{fig:ui-sparse}, we build user clusters where each cluster can be considered as a user cohort. For each cluster representing users with certain type of interest, we only consider a small subset of items to explore since it is not worth exploring many unrelated items in the corpus. For each user, we assign them to multiple top clusters and use the cluster weights when deciding how to connect user to the items in their corresponding clusters. The edges in the sparsified graph can be further converted to bandit parameters learnt online. Besides the intuitions, we provide a principled bandit framing in Section \ref{sec:diag-linucb}.

\textbf{Embedding-based graph construction.} With the graph view, now we describe how the graph is constructed offline. The idea is to leverage neural modeling. We first train a two-tower model to co-embed users and items, similar to the way many batch retrieval models \cite{yi2019samplingbias} are built nowadays.
Specifically, we train two neural networks, denoted by $f$ and $g$, to encode user and item features $\rvx, \rvy$ to embeddings $f(\rvx)$ and $g(\rvy)$ respectively. Given a batch of positive user-item pairs $\{\rvx_i, \rvy_i\}_{i=1}^B$, we use the batch softmax loss to train the two-tower model:
\begin{equation} \label{eq:two-tower}
\mathcal{L}(f, g) = -\sum_{i \in [B]}\log \frac{\exp{(\langle f(\rvx_i), g(\rvy_i)\rangle/\tau)}} {\sum_{j \in [B]}\exp{(\langle f(\rvx_i), g(\rvy_j)\rangle/\tau)}},
\end{equation}
where $\tau$ is a hyperparameter representing softmax temperature. 
In practice, using normalized embeddings, i.e., $\|f(\rvx_i)\|_2 = \|g(\rvy_i)\|_2 = 1$, can greatly improve trainability. As a result, we need to scale the logits to obtain meaningful softmax probabilities using $\tau$, as the logits are limited to the range of -1 to 1. 
We omit further details here and refer interested readers to \cite{yi2019samplingbias}, as the offline modeling is not the main focus of this paper.

Next, we apply off-the-shelf clustering algorithms to discretize the embedding space into a set of user clusters, based on a large sample of user embeddings.
For each user cluster, we choose the set of items with the highest embedding similarity measured by dot product between cluster centroid embeddings and item embeddings. This step largely narrows down the exploration space, allowing the online learning algorithm to focus on the items that have a higher probability of success.
More details can be found in Algorithm \ref{alg:graph-construction}. Note that it is possible to apply various clustering algorithms in our proposed framework, though we adopt kMeans in our system for simplicity. 

\begin{algorithm}[H]
\caption{Sparse Graph Construction }
\label{alg:graph-construction}
\begin{algorithmic}[1]
\STATE \textbf{Input}: A sample of user embeddings $\{\rvu_i\}_{i=1}^M$, item embeddings $\{\rvv_j\}_{j=1}^N$ as the target corpus. Here $\rvu_i$ and $\rvv_j$ are provided by a two-tower model trained according to Eq. \eqref{eq:two-tower}. Target number of items per cluster $W$.
\STATE Run a clustering algorithm (e.g., kMeans) on $\{\rvu_i\}_{i=1}^M$ to obtain C centroid embeddings $\{\rvc_c\}_{c=1}^C$. 
\STATE $\mathcal{I}_c \leftarrow \emptyset$ for all $c \in [C]$. $\mathcal{I}_c$ denotes the item set per cluster.
\FOR{c = 1 to C}
\STATE $\mathcal{I}_c \leftarrow $ \text{top-W items with largest values in set} $\{\langle \rvc_c, \rvv_j \rangle\}_{j \in [N]}$.
\ENDFOR
\STATE \textbf{Output}: $\{\mathcal{I}_c\}_{c=1}^C$.
\end{algorithmic}
\end{algorithm}

\subsection{Sparse Linear Bandits and Diagonal LinUCB} \label{sec:diag-linucb}
Now that we have explained the graph sparsification, we can proceed to introduce our bandit learning algorithm. We should note that if we assign each user to only one cluster during online learning, we can view the items ($\mathcal{I}_c$) in each cluster as distinct arms in a multi-armed bandit problem and utilize the UCB algorithm. Nonetheless, limiting each user to a single cluster could lead to a considerable loss of information from user embeddings. To tackle this issue, we instead assign each user to the closest $K$ (e.g., 10) clusters and also incorporate the cluster weights as part of the context representation. The question now becomes how to design an effective exploration-exploitation strategy to handle the many-to-many mapping from both users to clusters and items to clusters.
We propose a principled framing called \textsl{sparse linear bandits}, where the cluster weights can be effectively treated as a sparse context representation of a user in the high-dimensional space $\mathbb{R}^C$.

\textbf{Sparse linear bandits.} 
Given $C$ clusters, let $\rvw_u \in \mathbb{R}^C$ represent the weights of the top-$K$ clusters for the query from user $u$. Note that $\rvw_u$ is very sparse because $C$ is large and $\|\rvw_u\|_0 = K << C$. 
On the other hand, suppose for each item $j$, we have the ``ground-truth'' parameter $\vtheta_j^* \in \mathbb{R}^C$, where the $c$-th coordinate $\theta_{j,c}^*$ represents the quality or value of item $j$ for the cluster $c$. 
Based on the sparse graph in Fig. \eqref{fig:ui-sparse}, we can assume that $\theta_{j,c}^* = 0$ if there is no edge between item $j$ and cluster $c$. Accordingly, $\vtheta_{j}^*$ is also sparse, and we have $\|\vtheta_{j}^*\|_0 << C$ for most items. It is possible that some items (e.g., the popular ones with large audience) can belong to many clusters, but we could always control the sparsity of $\vtheta_{j}^*$ by setting a maximum degree per item. Similar to linear bandits, the reward $r_{u,j}$ is assumed to satisfy $\mathbb{E}( r_{u,j} | \rvw_u) = <\rvw_u, \vtheta_j^*>$. Based on our sparsity assumption, we can see that $\mathbb{E}(r_{u,j})$ is 0 for most $(u, j)$ pairs, and such pairs won't be explored in our algorithm. In other words, the \textsl{Large Action Space} problem of LinUCB is largely mitigated by our graph sparsification. Now we introduce a novel approximation, Diagonal LinUCB, to avoid computing the expensive $\textsl{Covariance Inversions}$ and to address the \textsl{Synchronous Updates} problem in the meantime. 

\textbf{Diagonal LinUCB.} The key idea is to only maintain and utilize the diagonal terms of covariance matrix $\rmA_{j,t}$ for item $j$ at step $t$. This is inspired by the momentum-based gradient descent methods (e.g., Adagrad and Adam) where only diagonal terms of Hessian matrix are used. Actually, covariance matrix $\rmA_{j,t}$ is essentially the Hessian matrix for solving linear regression. 
% From now on, we drop the subscript $t$ to ease the notations. 
From this point forward, we will omit the subscript $t$ to simplify the notation.
Let vector $\rvd_{j}$ denote the diagonal terms of $\rmA_{j}$ at some step. Let $d_{j,c}$ be the $c$-th coordinate of $\rvd_{j}$, and let $w_{u,c}$ be the $c$-th coordinate of $\rvw_u$. For a vector $\rvx$, let $\|\rvx\|_{supp}$ denote its support, i.e., the set of indices with non-zero entries. Then the update rules of $\rvd_j$ and $\rvb_j$ become:
\begin{align} \label{eq:diag_linucb_update}
d_{j,c} & \leftarrow d_{j,c} + w_{u,c}^2, \notag \\
% b_{j,c} & \leftarrow b_{j,c} + w_{u,c} \cdot r_{u,j},~ \text{for any}~ c~ \text{if}~ j \in \mathcal{I}_c,
b_{j,c} & \leftarrow b_{j,c} + w_{u,c} \cdot r_{u,j},~ \text{if}~ j \in \mathcal{I}_c (\forall{c} \in [C]),
\end{align}
where $\mathcal{I}_c$ is from Algorithm \ref{alg:graph-construction}. Furthermore, $UCB_j$ becomes
\begin{equation} \label{eq:ucbj}
UCB_j =  \sum_{c \in \|\rvw_{u}\|_{supp}} w_{u,c} b_{j, c} / d_{j, c} + \alpha \cdot \sqrt{\sum_{c \in \|\rvw_{u}\|_{supp}} w_{u,c}^2 / d_{j,c}},
\end{equation}
In exploitation mode, we drop the confidence bound term, and the estimated reward becomes
\begin{equation} \label{eq:diag-exploit}
    \hat{r}_{u,j} = \sum_{c \in \|\rvw_{u}\|_{supp}} w_{u,c} b_{j, c} / d_{j, c}.
\end{equation}
In the experiment section below, we will discuss how the exploitation mode is used in the use case of fresh content discovery. Putting things together, our detailed algorithm is shown in Algorithm \ref{alg:diag-linucb}.
\begin{algorithm}[H]
\caption{Diag-LinUCB Algorithm}
\label{alg:diag-linucb}
\begin{algorithmic}[1]
\STATE \textbf{Input}: Sparse graph $\{\mathcal{I}_c\}_{c=1}^C$ created by Algorithm \ref{alg:graph-construction}, context vector $\rvw_{u} \in \mathbb{R}^C$ at certain step from user $u$ by cluster assignment, number of clusters per user query $K$, number of items $N$,  hyperparameter $\alpha$,
\STATE Initialize $\rvd_{j} = \mathbf{I}, \rvb_j = \mathbf{0}$ for all $j \in [N]$.
\FOR{t = 1 to T}
\STATE Observe context vector $\rvw_u \in \mathbb{R}^C$.
% \STATE Set of triggered items $\hat{\mathcal{C}} \leftarrow \bigcup_{c \in \|\rvw_u\|_{supp}}\mathcal{I}_c$.
\STATE Identify the set of triggered items $\hat{\mathcal{C}} \leftarrow \bigcup_{c \in \|\rvw_u\|_{supp}}\mathcal{I}_c$.
\FOR{$j \in \hat{\mathcal{C}}$}
\STATE Compute $UCB_j$ according to Eq. \eqref{eq:ucbj}.
\ENDFOR
\STATE Select item $a = \arg\max_{j \in \hat{\mathcal{C}}} UCB_j$. 
\STATE Observe reward $r_{u,a}$.
\STATE Update $\rvd_a$ and $\rvb_a$ using Eq. \eqref{eq:diag_linucb_update}.
\ENDFOR
\end{algorithmic}
\end{algorithm}

\textbf{Discussion on Eq. \eqref{eq:diag_linucb_update}}. Now we take a closer look at the parameter updates in Eq. \eqref{eq:diag_linucb_update}. Given a reward $r_{u,j}$, the update of $b_{j,c}$ gives higher weights to clusters closer to the user embedding. In other words, for user that is more familiar with an explored item, their feedback on the item will be trusted more. It is worth noting that when $w_{u,c} = 1$, Eq. $\eqref{eq:diag_linucb_update}$ is reduced to multiple $UCB$ that run separately for each cluster. As shown in experiments, we find that incorporating cluster weights can outperform treating all clusters equally. Compared to LinUCB, the updates in Diag-LinUCB are much more light-weighted, and more importantly, do not require the item-level synchronization since the updates are fully distributed over the edges in the sparse graph. This property eases the design of online learning infrastructure and significantly improves the system throughput.  

\textbf{Context vector}. There could be a few options for computing the context vector $\rvw_u$ from user embedding $\rvu$. The naive way is to let $w_{u,c} = \langle \rvu, \rvc_c\rangle$ where $\rvc_c$ is the embedding of each centroid. 
As mentioned in Section \ref{sec:sparse_bipartite_graph}, we apply embedding normalization and softmax temperature when learning the two-tower model, which usually leads to top clusters having similar weights, with small numerical differences.
This is expected because small difference in logits is amplified by the temperature $\tau << 1$ and the softmax function during training. Directly using the logits as context vector does not reflect the true user preferences over various clusters.
To address this issue, one option is to use a softmax transform similar to the one used in the training stage, namely
\begin{equation} \label{eq:context-vector}
w_{u,c} = \frac{\exp(\langle \rvu, \rvc_c\rangle/\tau')}{\sum_{c' \in [C]} \exp(\langle\rvu, \rvc_{c'}\rangle/\tau')},
\end{equation}
where $\tau'$ is another hyperparameter. One limitation of this approach is that we need to empirically tune $\tau'$. Another option is to train a multi-class classification model to directly predict the distribution over user clusters. However, this requires a much more intricate pipeline that involves training both the two-tower model and the user clusters. In this paper, we choose the simple approach in Eq. \eqref{eq:context-vector} and leave the second option to future work.

%% file: system_overview.tex
\section{System Overview}
In this section, we provide an overview of the Online Matching system. We introduce the end-to-end workflow, followed by a detailed overview of the online agent in Section \ref{sec:online_agent}.

\subsection{End-to-end workflow}
\label{sec:workflow}
The entire Online Matching workflow consists of an offline pipeline and an online agent responsible for the closed-loop learning, as shown in Fig. \ref{fig:e2e}. The offline pipeline produces a sparse bipartite graph, as introduced in Section \ref{sec:sparse_bipartite_graph}, which is adopted by the online agent. It has the following components:
\begin{figure*}[ht]
  \centering
  \includegraphics[width=0.75\textwidth]{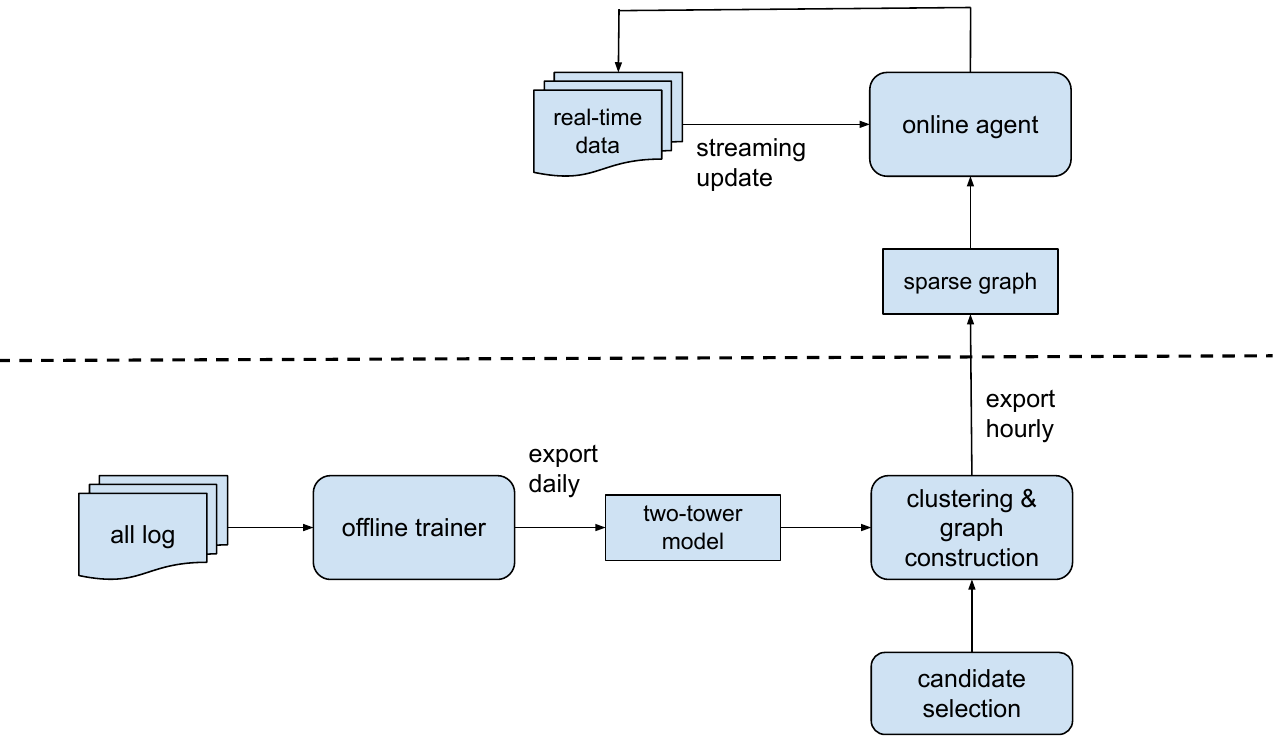}
  \caption{An overview of the end-to-end Online Matching system. Components below the dashed line are the offline pipeline responsible for generating the sparse bipartite graph. Components above the dashed line represent the online system responsible for real-time learning in a closed loop.}
  \label{fig:e2e}
  \Description[A diagram of the end-to-end Online Matching system.]{The offline pipeline consists of an offline trainer which exports a two-tower model daily, and a clustering \& graph construction module that exports sparse graphs hourly. The online system takes sparse graphs generated by the offline pipeline and uses an online agent to learn from real-time user feedback.}
\end{figure*}

\begin{itemize}

\item \textbf{Two-tower model trainer.}
We train an offline two-tower model by sequentially consuming a large amount of logged user feedback over time. The sequential training ensures that the model can adapt to the distribution change in the latest batch of data.
As mentioned earlier, this model encodes item features, allowing it to generate meaningful embeddings even for newly added items. The two-tower model is exported on a daily basis, with both towers used by the downstream components to create the sparse bipartite graph. The user tower is also used by the online system to generate user embedding $\rvu$ and context vector $\rvw_u$.

\item \textbf{Candidate selection.} 
This component creates a corpus of items eligible for exploration. Multiple filters are applied to ensure the selected candidates satisfy our strict trust-and-safety criteria.
In this paper, our system is mainly focused on exploring fresh videos, therefore a rolling time window that covers a few days is used for item selection. We also apply various quality thresholds to balance the quality and size of the corpus. 

\item \textbf{Clustering and graph building.}
Once clustering is finished based on the two-tower model exported most recently, graph builder is triggered to build the sparse graph according to Algorithm ~\ref{alg:graph-construction}. 
The graph building process is executed in both batch and real-time modes concurrently. In batch mode, graph builder takes the output of the candidate selection step, and exports a new graph every few hours. Real-time mode complements batch mode by incrementally updating the sparse graph with newly eligible items to ensure a small latency for items to enter the exploration pool. 
\end{itemize}

Online agent represents the system that conducts the bandit algorithm and aggregates user feedback in real-time. It takes the sparse graph produced from the above pipeline as input. Whenever the sparse graph is updated, bandit parameters are synchronized in online agent with low latency: 
% new edges are added with infinite confidence bound 
new edges are added with infinite confidence bound (so that they will be prioritized for future exploration),
and old edges that are only in the previous graph version are removed. In the next section, we will delve into the specifics of online agent.

\subsection{Online Agent} \label{sec:online_agent}

\begin{figure*}[ht]
  \centering
  \includegraphics[width=0.8\textwidth]{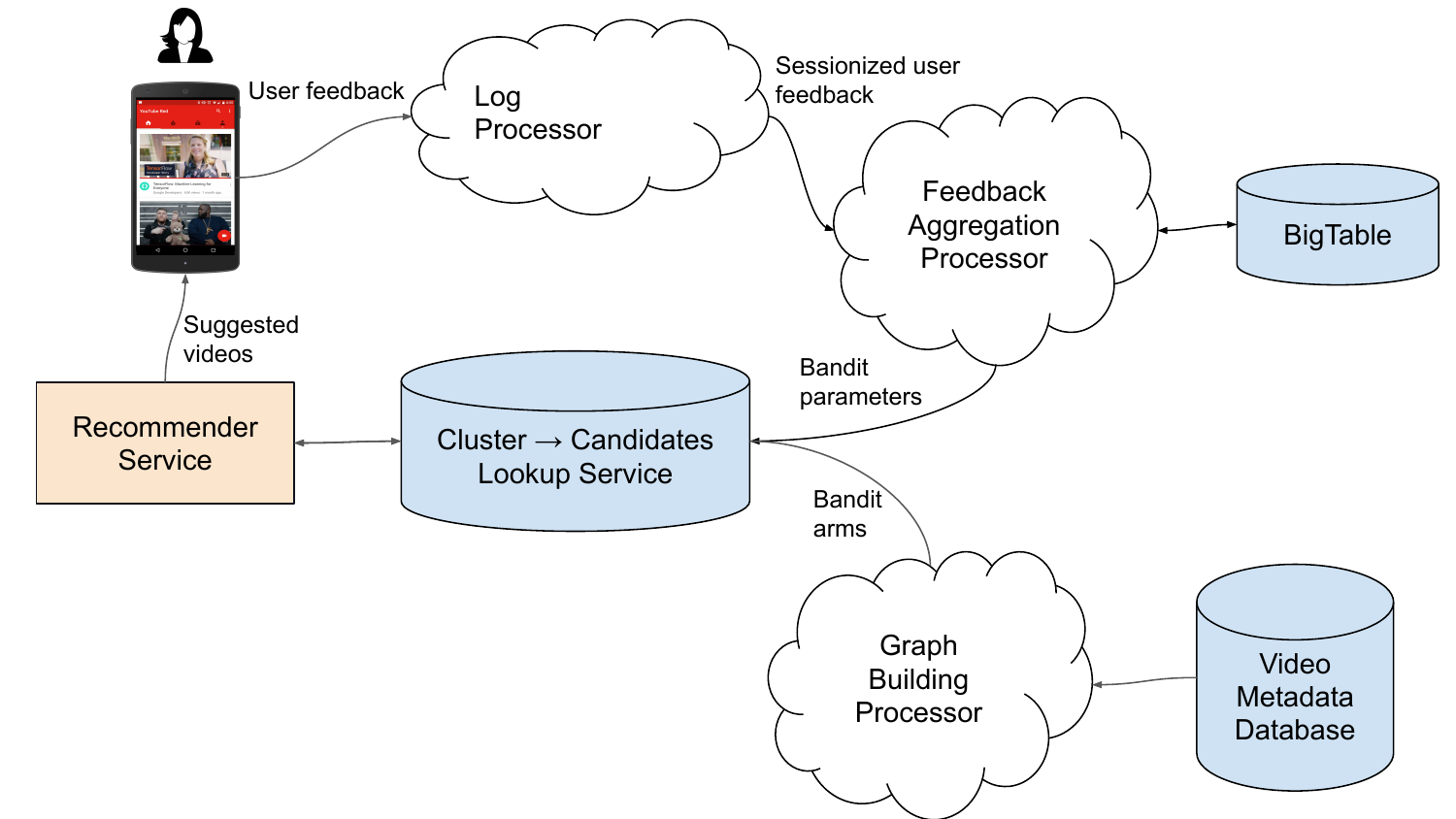}
  \caption{A closer look at how the online agent aggregates users' feedback and accommodates newly eligible videos.}
  \label{fig:feedback}
  \Description[A diagram of Online Matching aggregating users' feedback and accommodating newly eligible videos.]{User feedback flows through log processor, feedback aggregation processor (which exchanges information with BigTable), cluster to candidate lookup service, and finally recommender service. Graph building processor monitors the video metadata database and updates the cluster to candidates lookup service with newly eligible videos.}
\end{figure*}

As shown in Fig. \ref{fig:feedback}, the key components of online agent are chained as a closed loop. 
Explored items from Online Matching are allowed to be shown at a fixed position in the UI, so that users' direct feedback on them can be measured without being affected by 
% ranking policies.
existing ranking policies. 
The log processor is used to incrementally generate various kinds of engagement signals on the explored items in the format compatible with the downstream jobs. The feedback aggregation processor is built on top of Bigtable ~\cite{bigtable-osdi06} and is responsible for aggregating pair-wise (cluster and video) bandit parameters according to Eq. \eqref{eq:diag_linucb_update}. As illustrated in Table ~\ref{tbl:toy-example}, each row in the Bigtable represents one cluster, and each column represents the corresponding items of that cluster in the sparse graph. Conceptually, Bigtable is a sparsely populated table that can scale to the billions of rows and columns, and is compatible with the proposed Diag-LinUCB algorithm.

\begin{table}[h]
  \begin{tabular}{|c|c| }
    \hline
     \textbf{Column} & \textbf{Cell value} \\
    \hline
     feedback:$item_1$ & \{ $d_{1,1}$: $54.4$, $b_{1,1}$: $624.2$, $w_{1,1}^{2}$: $1.5$ \} \\
    \hline
     feedback:$item_2$ & \{ $d_{2,1}$: 57.6, $b_{2,1}$: 144.6, $w_{2,1}^{2}$: 1.8 \} \\
    \hline
     feedback:$item_3$ & \{ $d_{3,1}$: 76, $b_{3,1}$: 547.1, $w_{3,1}^{2}$: 2.6 \} \\
    \hline
  \end{tabular}
  \caption{Illustration of how the aggregated bandit parameters $d_{j,c}$, $b_{j,c}$, $w_{j,c}^{2}$ in Eq. \eqref{eq:diag_linucb_update} are stored in Bigtable.  Row keys correspond to hashed cluster IDs. Here we demonstrate one row corresponding to cluster 1 and its three columns.} 
  \label{tbl:toy-example}
\end{table}

Bandit parameters in Bigtable are frequently pushed to the cluster-to-candidates lookup service. The recommender service is used for obtaining user clusters, and look up the candidates and their bandit parameters from the lookup service. The recommender service is further used to rank all the candidates according to the UCB in Eq. \eqref{eq:ucbj} or Eq. \eqref{eq:diag-exploit} if the goal is to exploit high quality candidates.

\begin{figure}[ht]
  \centering
  \includegraphics[width=0.5\textwidth]{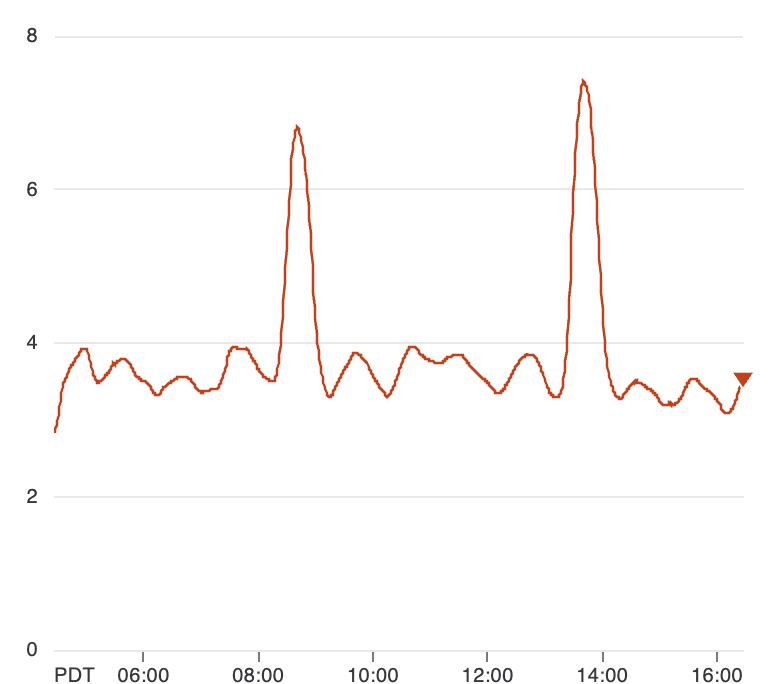}
  \caption{A plot of the number of candidates with infinite scores over time, due to the addition of new eligible items.
    The two peaks correspond to the moments when the batch graph builder injects a new batch of items into the bipartite graph.
}
  \label{fig:inf-score}
  \Description[A plot of the number of candidates with infinite scores over time.]{X-axis is time and y-axis is the number of candidates with infinite scores. There are two peaks in the curve.}
\end{figure}

It is worth noting that, compared to the classic bandits setup with a fixed set of arms, we always have fresh items added as new arms in Online Matching. In particular, fresh items are \textsl{continuously} injected into the bipartite graph through the graph building pipeline. New items that have never been explored would have an infinite confidence bound in Eq. ~\eqref{eq:ucbj} and are therefore prioritized for exploration in the recommender service. Due to the batch addition of new items, we can observe spikes of infinite UCB scores as shown in Fig. \ref{fig:inf-score}. The spikes usually disappear quickly, demonstrating that users' feedback to new items are quickly incorporated in bandit parameters.

\subsection{System Performance}
To demonstrate the real-time performance of Online Matching, we measure the following two types of update latency:
\begin{itemize}
    \item \textsl{Policy update latency}: This is the period of time from the point user sees the explored item, to the point when user's feedback on the item is incorporated in bandit parameters contained in the lookup service in Fig. \ref{fig:feedback}. 
    Since our primary application is video recommendation, this latency also includes user's watch time on video capped at a certain value, as a major part of the latency in the log processor. Actually sessionizing user feedback in the log processor contributes to most of the policy update latency.
    \item \textsl{Corpus update latency}: This is the period of time from the point a fresh item is eligible for exploration to the point when the item is added to the sparse graph. 
\end{itemize}
The median and the 95-th percentile of both latency are summarized in Table \ref{tbl:latency}. Besides latency, our system achieves high throughput, e.g., it can handle millions of bandit updates per second, allowing it to scale to billions of users.

\begin{table*}
  \begin{tabular}{ |c|c|c|c| }
    \hline
     & P50 (minutes) & P95 (minutes) & Throughput (updates/second)\\
    \hline
    Policy update latency & 45 & 74 & O(1M) \\
    \hline
    Corpus update latency & 41.1 & 60.1 & O(1K) \\
    \hline
  \end{tabular}
  \caption{Policy update latency, corpus update latency and the system throughput.}
  \label{tbl:latency}
\end{table*}

Particularly, the low latency of policy update is critical for recommendation quality since the expected regret grows as the feedback delay increases \cite{delayfeedback-JoulaniGS13}. To empirically verify the expected regrets, we add artificial latency into the aggregation processor in Fig. \ref{fig:feedback}. As shown in Table ~\ref{tbl:latency-inject}, as latency was introduced, the agent became less capable of identifying low-performing bandits, resulting in a decrease in CTR and total rewards on explored items.

\begin{table}[h]
  \begin{tabular}{ |c|c|c| }
    \hline
     &  CTR & Total Rewards\\
    \hline
    Baseline with no artificial delay injected & - & - \\
    \hline
    20 min delay added & -2.82\% & -11.82\% \\
    \hline
    40 min delay added & -4.4\% & -22.84\% \\
    \hline
  \end{tabular}
\caption{A study of artificial latency injection in policy updates' impact on CTR (click-through rate) and total rewards (measured by multiple user satisfaction and engagement metrics).}
  \label{tbl:latency-inject}
\end{table}

%% file: results.tex
\section{Experiment Results}
We conduct our experiments on one major recommendation surface of YouTube -- one of the world’s largest video content discovery, creation and sharing platforms. Similar to many industry-scale recommender systems, the recommendation surface we considered consists of multiple stages including candidate generation and a multi-task ranking system \cite{zhe19watchnext}. Given a user and a corresponding context, the candidate generation stage employs multiple video retrieval systems to generate a few hundred/thousand candidate videos with a primary focus of optimizing the recall performance. The ranking stage then takes the set of retrieved videos as input and generates the final ranking. In the following two use cases, Online Matching is added as a new retrieval source to the candidate generation stage.

\subsection{Use Cases}
We primarily consider two use cases of Online Matching:
\begin{itemize}
\item \textsl{Fresh Content Discovery (Type-I).} Content creators upload millions of videos to Youtube on a daily basis. These videos vary a lot in quality and traditional batch-learning based recommender systems are limited in their capability to identify high-quality fresh videos and recommend them to the appropriate audience in a timely manner.
Intuitively, an efficient exploration system on fresh content can supplement traditional recommender systems by identifying and leveraging quality fresh content to improve user experience with satisfied engagement, particularly with fresh content. To compensate the quality loss from exploration, the idea is to use a very small traffic for exploration, and conduct exploitation (i.e., amplifying high-quality candidates) for the rest of traffic.
\item \textsl{Corpus Exploration (Type-II).} Real-world recommender systems typically have a long-tail distribution over the recommendation corpus. Traditional recommender systems are primarily exploitation-based due to a feedback loop that reinforces the existing ``winners'' in the corpus. Exploring and leveraging the long-tail portion of the corpus is both challenging given its vast size and rewarding. In this case, the goal is to grow the \textsl{discoverable corpus}. Having a larger discoverable corpus is argued to bring long-term value to recommenders, e.g., through reducing system uncertainty on sparse data region \cite{Chen2021}. It can also provide torso or long-tailed content creators more opportunities to showcase their content to a wider audience. However, similar to Type-I, explicitly exploring long-tail and fresh items can lead to short-term loss of user engagement and satisfaction. Therefore, the goal of this use case is to enlarge discoverable corpus while having minimal regret on short-term engagement.
\end{itemize}

Online Matching is a fitting solution for the above use cases thanks to its efficiency in both the exploration algorithm and system. On one hand, Online Matching trims the exploration space through offline learning, making it possible to only use small traffic for exploration in the first use case. On the other hand, the real-time property of Online Matching system helps to minimize exploration errors, so that it can effectively enlarge the discoverable corpus while maintaining a small regret on short-term engagement in the second use case.

\subsection{Fresh Content Discovery}
\subsubsection{Experiment Setup.}

We select up to $O(1M)$ fresh videos that were uploaded within the past certain days (referred as X days in the rest of this paper) as the exploration corpus. Multiple offline and online filters are applied to make sure the explored videos are safe and satisfy minimum quality requirements. We use a small size of frequently shuffled user traffic ($\leq 2\%$) to explore this corpus and learn the bandit parameters. In the exploration mode, for getting unbiased user feedback, the picked candidate from Online Matching bypasses the ranking layer and is shown at a fixed position in the UI. We call one experiment running in this exploration mode as \textsl{one exploration slot}. The reward we use is a combination of multiple signals representing user's happiness with the recommended videos. If any video is filtered during exploration, the filtering information is also used as part of the reward in bandits. It's worth noting that, besides hard filtering, the sparse graph construction can also control candidate quality through the offline learnt embeddings.

To amplify the learnt good fresh videos to all users, we also set up another online agent that works in an ``exploitation'' mode for the rest of traffic (98-99\%). Particularly, it reuses the corpus and bandit parameters from the aforementioned online agent and ranks videos only based on the estimated reward in Eq. \eqref{eq:diag-exploit}. In exploitation, since there is no need to collect user feedback, multiple top candidates by mean reward are passed to the ranking layer together with candidates from other sources. Exploration corpus is being updated on a rolling basis -– new eligible videos are continuously injected to the exploration corpus and videos uploaded beyond X days continuously graduate from the corpus.

\textbf{Top-k randomization.} Rather than getting feedback instantly in the theoretical bandit model, our system still has a nontrivial policy update latency. This latency can cause the system to overly explore certain items before collecting their reward. To address this issue, we introduce a randomization mechanism to uniformly sample a video from the top $k$ videos measured by UCB. We set $k = 5$ in the following experiments.

\subsubsection{Results.}
We ran an online user-diverted A/B testing for a week to understand the effectiveness of our \textsl{explore-and-amplify} framework.
For both exploration and exploitation modes, the control is the production recommender system without adding Online Matching agent as an additional candidate generator.

\begin{figure}[ht]
  \centering
  \includegraphics[width=0.45\textwidth]{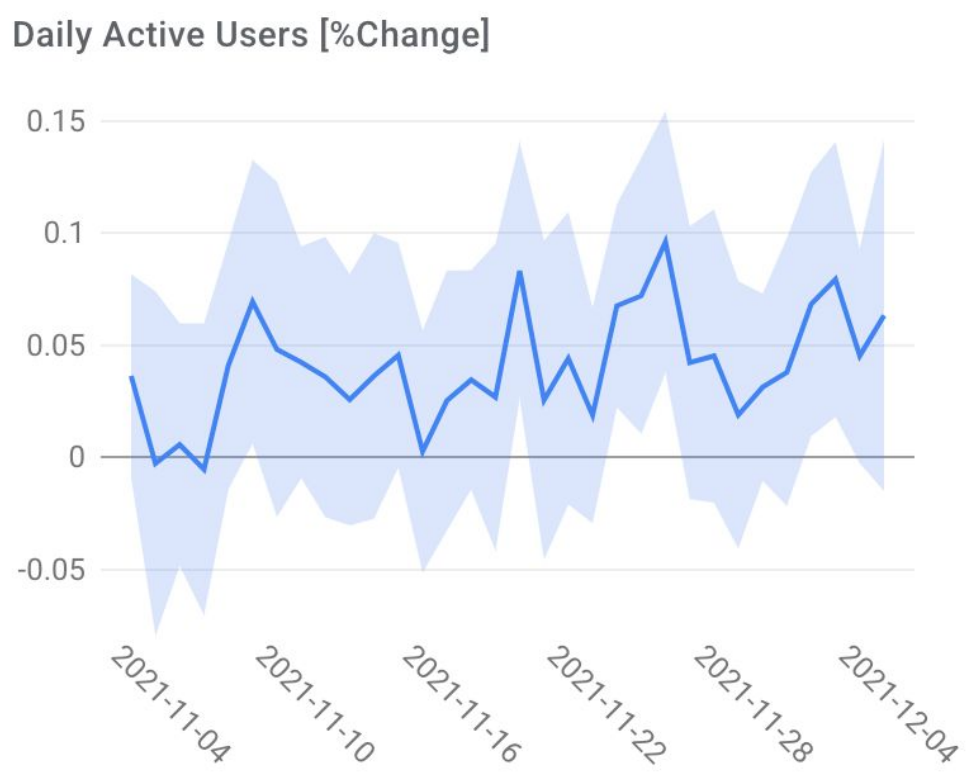}
  \caption{Holdback result of Online Matching exploitation mode for Fresh Content Discovery.}
  \label{fig:dau}
  \Description[A plot of the Daily Active User percentage change curve over a month.]{X-axis is time and y-axis is the Daily Active Users percentage change. There is a slight up-trend over the one month period.}
\end{figure}

\textbf{Gains from exploitation}. The results of Online Matching exploitation mode are shown in Table \ref{tbl:freshcontentdiscovery}. We use the setup where all user clusters are treated equally as the baseline to show the value of our user context modeling. We report two Diag-LinUCB arms where the second one uses a larger corpus and more clusters. Note that with the larger graph, we had to increase exploration traffic from 1\% to 2\% to avoid under exploration. Overall, we see improved topline satisfied engagement metric from Online Matching and significant gains on the fresh item slice.
To study long-term impact, we have run a holdback for several weeks. As demonstrated in Fig. \ref{fig:dau} shows, there is a +0.04\% improvement on daily active users, a long-term metric that is much more difficult to improve than short-term user engagement.

\begin{table*}
  \begin{tabular}{ |c|c|c|c|c|c|c| }
    \hline
     & \shortstack{Satisfied \\ user engagement} & \shortstack{Engagement with \\ fresh content} & \# clusters & Fresh corpus size & \shortstack{Graph size \\ (\# edges)}\\
    \hline
    Equal-weight Bandit & +0.03\% & +3.61\% & 15k & 1x & $\sim$ 4M \\ \hline
    Diag-LinUCB         &+0.08\%  & +5.25\% & 15k & 1x & $\sim$ 4M \\ \hline
    Diag-LinUCB (Larger Graph)  &+0.15\%  &+8.33\% & 30k & 3x & $\sim$ 20M \\ \hline  
  \end{tabular}
  \captionof{table}{Fresh Content Discovery exploitation A/B testing results. Improvements over the production system without Online Matching are reported. The first two arms used $1\%$ traffic for exploration, and the 3rd arm used $2\%$ traffic for exploration due to graph being larger.}
  \label{tbl:freshcontentdiscovery}
\end{table*}

\textbf{Cost of exploration}. We observed -0.16\% and -0.19\% satisfied user engagement loss from the 1\% and 2\% exploration slots for the 2nd and 3rd rows in Table \ref{tbl:freshcontentdiscovery}. By discounting the engagement loss with the small traffic proportion, it is clear that the value added by exploitation far outweighs the cost of exploration.

\begin{figure*}[ht]
  \centering
  \includegraphics[width=0.8\textwidth]{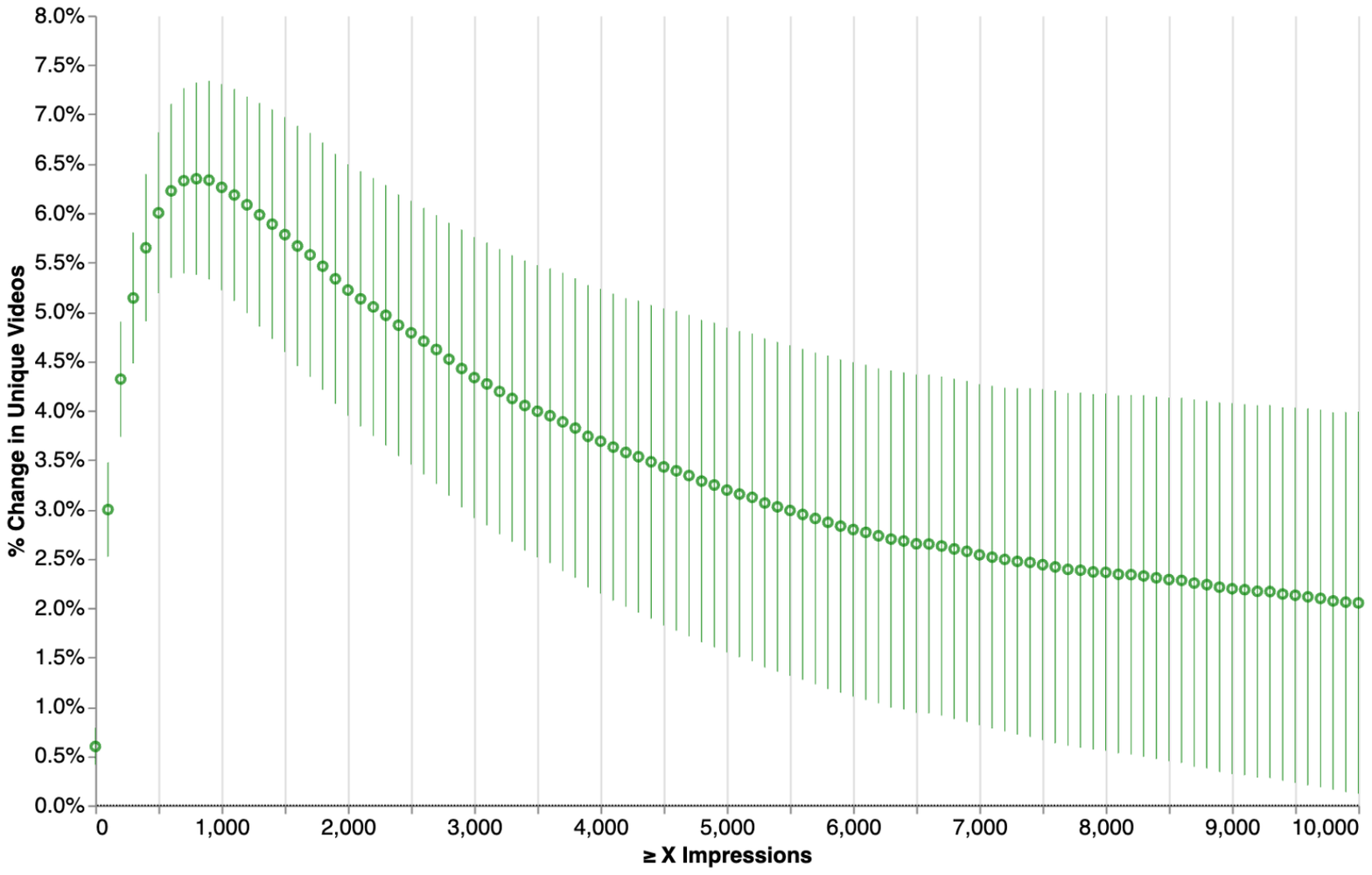}
  \caption{Relative change in terms of the Daily Discoverable Corpus Size metric across different impression thresholds.}
  \label{fig:duiv}
  \Description[A plot of the Daily Discoverable Corpus Size percentage change curve over different impression thresholds.]{X-axis is impression threshold and y-axis is percentage change in unique videos. Improvement is observed across all the impression thresholds with a  peak around the threshold of 1000.}
\end{figure*}
\subsection{Corpus Exploration}
\subsubsection{Experiment Setup}
In the traditional A/B testing setup, user traffic is randomly assigned to a control and treatment group. This user-diverted setup can enable measuring user-side metric changes. In this use case, we want to measure the growth of discoverable corpus. The user-diverted experiment is not an appropriate method for measuring corpus changes because the control and treatment group share the same corpus. As a result, any treatment effect on the corpus can be leaked to the control group. In order to correctly measure corpus changes, we employ a user-corpus co-diverted setup where we partition the entire corpus (e.g., by hashing item id) into multiple disjoint slices and expose each slice to a faction of user traffic in one experiment.

To largely explore the long-tailed proportion of our corpus, we curate a large corpus of fresh videos,  that is $O(10M)$ videos uploaded within X days. We divide the entire corpus into 10 slices and one slice in each experiment that has $6\%$ user traffic. Type-II experiment shares many similar configurations such as the two-tower model and number of clusters as Type-I, except that the exploration corpus is much larger
and the online agent only runs in the \textsl{exploration} mode in this use case.

\subsubsection{Experiment Results}

We compare daily unique videos greater than different impression threshold values to measure the discoverable corpus sizes across control and treatment groups, each of which contains 1/10 of the entire fresh corpus.
Fig. \ref{fig:duiv} shows the relative corpus size improvement across multiple impression values. Note that we can naively boost the impressions of more unique videos regardless of their performance. But overly showing low-quality videos to users can cause very negative user experience. Overall, we only observed -0.05\% topline user engagement loss with corpus size gain from Fig. \ref{fig:duiv}. Comparing to Type-I, here the cost of the exploration is much lower. 

This is expected because in one exploration slot, Type-II has a slightly smaller exploration corpus than Type-I, but its exploration user traffic is significantly larger ($6\%$ vs $2\%$). Intuitively, this means that in Type-II bandits can discover high-quality items more quickly.

%% file: conclusion.tex
\section{Conclusion}

In this paper, we introduced Online Matching, a large-scale real-time bandit system for item exploration and recommendation.
To scale up to serving billions of users and millions of items, we proposed an algorithm-system co-design that combines offline and online learning, introduces a novel Diagonal LinUCB algorithm, and devises a high-throughput online agent system. Empirically, we applied Online Matching to two important use cases in YouTube and demonstrated its effectiveness for increasing fresh content adoption and growing discoverable corpus.

We hope our exercise of building a real-world bandit system can provide some valuable insights for future endeavors to address practical challenges. For example, we used a mixture of exploration and exploitation modes for our Fresh Content Discovery use case. In contrast to the classic bandit setup where minimizing regret in a single mode is the primary goal, more research needs to be done to study the mixture setup. In addition, real-world applications often have to deal with a dynamic exploration corpus, where items are being constantly added or removed in a streaming fashion. Designing principled bandit algorithms for dynamic arms remains an interesting open research problem.